\documentclass[11pt]{article}
\usepackage{acl}
\usepackage{times}
\usepackage{url}
\usepackage{latexsym}

\usepackage[utf8]{inputenc}
\usepackage{pmboxdraw}  

\usepackage{microtype}

\usepackage{inconsolata}

\usepackage{import}

\usepackage{url}
\usepackage{booktabs}
\usepackage{todonotes}
\usepackage{siunitx}
\usepackage{mathtools}
\usepackage{makecell}
\usepackage{adjustbox}
\usepackage{amsmath}
\usepackage{amsfonts, amssymb}

\usepackage{nccmath}    
\usepackage{pgf}
\usepackage{hyperref}
\usepackage{tikz}

\usetikzlibrary{arrows.meta, positioning, shadows, shapes}
\tikzstyle{arrow} = [thick,->,>=stealth]
\tikzstyle{N} = [draw, fill=white, minimum size=1cm, align=center, line width=.3mm]
\tikzstyle{dcs} = [draw, double copy shadow, shadow xshift=2pt, shadow yshift=-2pt]
\tikzstyle{cyl} = [cylinder,  draw = violet,  text = black, cylinder uses custom fill,  cylinder body fill = magenta!10,  cylinder end fill = magenta!40, aspect = 0.2,  shape border rotate = 90, align=center]
\tikzstyle{scs} = [copy shadow, shadow xshift=2pt, shadow yshift=-2pt]
\tikzstyle{startend} = [ellipse, minimum width=3cm, minimum height=1cm, text centered, draw=black, fill=white]
\tikzstyle{io} = [trapezium, trapezium left angle=70, trapezium right angle=110, minimum size=5mm, text centered, line width=.3mm, draw=black, fill=white]

\definecolor{orange}{RGB}{255,180,0}
\definecolor{blue}{RGB}{20, 100, 255}
\definecolor{pink}{RGB}{255, 50, 220}
\definecolor{green}{RGB}{0, 200, 0}
\definecolor{yellow}{RGB}{255, 255, 50}

\usepackage{pgfplots}
\usepackage{multirow}
\usepackage{graphicx}
\usepackage{color, colortbl}
\usepackage{tabularx}

\everymath=\expandafter{\the\everymath\displaystyle}

\usepackage{array}
\usepackage{makecell}
\usepackage{geometry}

\aclfinalcopy

\title{Resource-Efficient Language Models:\\Quantization for Fast and Accessible Inference}

\author{
Tollef Emil Jørgensen \\
Department of Computer Science \\
Norwegian University of Science and Technology \\
Trondheim, Norway \\
{\tt tollef.jorgensen@ntnu.no} \\
}
\date{}

\begin{document}
\maketitle

\begin{abstract}
Large language models have significantly advanced natural language processing, yet their heavy resource demands pose severe challenges regarding hardware accessibility and energy consumption.
This paper presents a focused and high-level review of post-training quantization (PTQ) techniques designed to optimize the inference efficiency of LLMs by the end-user, including details on various quantization schemes, granularities, and trade-offs. The aim is to provide a balanced overview between the theory and applications of post-training quantization.

\end{abstract}

\section{Introduction}
\label{sec:intro}
Modern language models have demonstrated remarkable capabilities across various natural language processing tasks, from text generation to complex reasoning and other \textit{emergent abilities} \citep{wei2022emergentabilitieslargelanguage}. However, through years of scaling, large language models (LLMs) have become incredibly resource-intensive, posing challenges for deployment, e.g., in environments with limited computational resources and applications where latency is prioritized. These factors hinder their accessibility and widespread adoption by end-users without relying on centralized APIs, due to the sheer size of the models.
The problem of reducing the cost of deep learning based models through limited numerical precision has been studied far before the era of LLMs \citep{gupta-deep-learning-2015, courbariaux2015trainingdeepneuralnetworks}, later commonly referred to as \textit{quantization} \citep{han2015deep, jacob2017quantizationtrainingneuralnetworks}.
A significant portion of efficiency issues in LLMs stem from the complexity of the transformer architecture \citep{vaswani2017attention}, which forms the backbone of most models, including the openly available ones such as the OPT- \citep{zhang2022optopenpretrainedtransformer}, OLMo- \citep{groeneveld2024olmoacceleratingsciencelanguage, olmo20252olmo2furious}, Llama- \citep{touvron2023llama, grattafiori2024llama}, Phi- \citep{gunasekar2023textbooks, abdin2024phi}, and Gemma-series of models \citep{gemmateam2024gemma2improvingopen, team2025gemma}.
Architectures like Mamba \citep{gu2024mambalineartimesequencemodeling} -- a structured state-space sequence model \citep{gu2022efficientlymodelinglongsequences}, RWKV \citep{peng2023rwkvreinventingrnnstransformer} -- speeding up inference with RNNs, and Hyena \citep{poli2023hyena} -- replacing Transformers' attention with efficient long convolutions, are also being researched to resolve some of the complexity issues. Drastic changes to the transformer architectures, such as replacing all computations with 1-bit weights (from matrix multiplications into pure additions), have also proven feasible \citep{wang2023bitnet, ma2024era}.

The complexity in transformers is connected to the self-attention mechanism, which results in robust handling of long-range dependencies, but as a result, scales quadratically to the input sequence length. The problem is compounded by the observed \textit{scaling laws} of language models \citep{kaplan2020scalinglawsneurallanguage}, which caused model developers to go above and beyond with parameter counts, solving performance issues with \textit{more compute}.

Moreover, not only is the hardware required to run these models at acceptable speeds an issue, but also the amount of memory required. While these constraints are linked, merely storing a 70 billion parameter model in single precision requires approximately 280 GB of available GPU memory -- at the cost of over \$100.000 with, e.g., 4x H100.

In the transformer architecture, matrix multiplcations residing in the attention
projection and feed-forward layers account for approximately 95\% of all parameters (for dense models of 6.7B parameters and above), corresponding to up to 85\% of compute, motivating the need for speed-ups in these operations \citep{dettmers-gptint8}, where quantization is one of many possible resolutions.
\\
\\
Studies related to efficient language models are described in Section \ref{sec:related}, followed by background material for quantization and in-depth details in Section \ref{sec:bg}
Section \ref{sec:app:quantization} provides an overview of post-training quantization methods.
Concluding remarks are found in Section \ref{sec:conclusion}.

\section{Related Work}
\label{sec:related}
While there are excellent works related to the topic of efficient inferencing already \citep{zhou_survey_2024, zhu_survey_2024, bai_beyond_2024, farina_sparsity_2024}, they typically span larger aspects, such as data- and system-level optimization in pre-training stages, evaluation practices, and smaller models in domain-specific applications \citep{wang_comprehensive_2024}. Others are focused on quantization during training \citep{chitsaz2024exploringquantizationefficientpretraining}.
Studies aligning more with quantization for efficient inferencing \citep{existing-nagel2021whitepaperneuralnetwork, existing-gholami2022survey} are quickly becoming outdated as methods have been developed incredibly quickly. They do, however, provide a great foundation for quantization research.
Some recent surveys, while technically detailed, lack comprehensive overviews \citep{existing-li2024contemporary, shen-exploring-llm-quant-2024}, or focus in great detail on a smaller subset of methods or models \citep{existing-lang2024comprehensive, jin2024comprehensiveevaluationquantizationstrategies}.
A highly detailed and recently updated survey on low-bit quantization by \citet{gong2024surveylowbitlargelanguage} includes many of the methods referred to in this work, although with a focus beyond that of PTQ for the end-user, aimed more towards researchers interested in a low-level description of the involved algorithms, alternative datatypes, and toolkits.

This work aims to provide a balanced introduction to the theory behind quantization, including content useful for future research and development of post-training model quantization, with the aim of increasing accessibility of larger models with regular hardware.

\section{Background}
\label{sec:bg}
Quantization refers to techniques that reduce the precision of the weights and activations of neural networks, resulting in smaller but also \textit{faster} models, due to optimizations for hardware acceleration.
Publishing quantized models has become a common way of distributing pre-trained models, especially since LLMs became more accessible after early model releases such as Llama \citep{touvron2023llamaopenefficientfoundation} and Mistral \citep{jiang2023mistral7b} with their modest sizes of 7B parameters and above.
Represented initially in high-precision formats (e.g., FP32), the components of a network should, through quantization, be mapped to a discrete set of values in a lower-bit format (e.g., INT8) while minimizing information loss, preserving as much of the original capabilities of a model.

\begin{figure}
    \centering
    \includegraphics[width=1.0\linewidth]{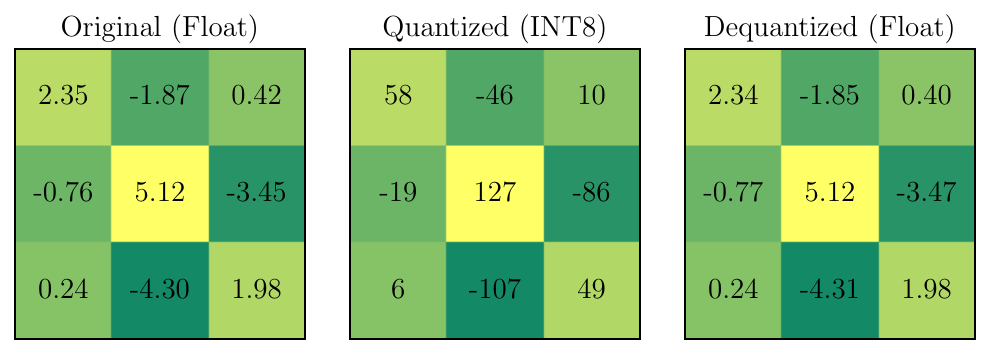}
    \caption{Quantization and dequantization of a $3\times3$  matrix visualized, with rounded values for readability. In the process from FP32 to INT8 and back to FP32, most values will deviate.}
    \label{fig:quantization-viz}
\end{figure}

Figure \ref{fig:quantization-viz} illustrates how arbitrary values may shift during quantization of values in FP32 to INT8 and back to FP32 for a $3\times3$ matrix.

Although quantization is no new phenomenon, it has been increasingly studied after we saw the size of neural networks grow throughout the late 2010s, becoming troublesome to deploy for resource-constrained applications \citep{existing-gholami2022survey}.
This effect has become more prevalent recently, specifically with \textit{open-weight} models, enabling researchers and hobbyists alike to download and use these models freely through services like Hugging Face\footnote{\url{https://huggingface.co/models}}  -- hosting more than 1.5 million models as of March 2025, with a large part of them being quantized versions, and plays a significant part in democratizing LLMs and their widespread adoption \citep{shashidhar2023democratizing}.

\begin{figure*}[!ht]
    \centering
    \includegraphics[width=1.0\linewidth]{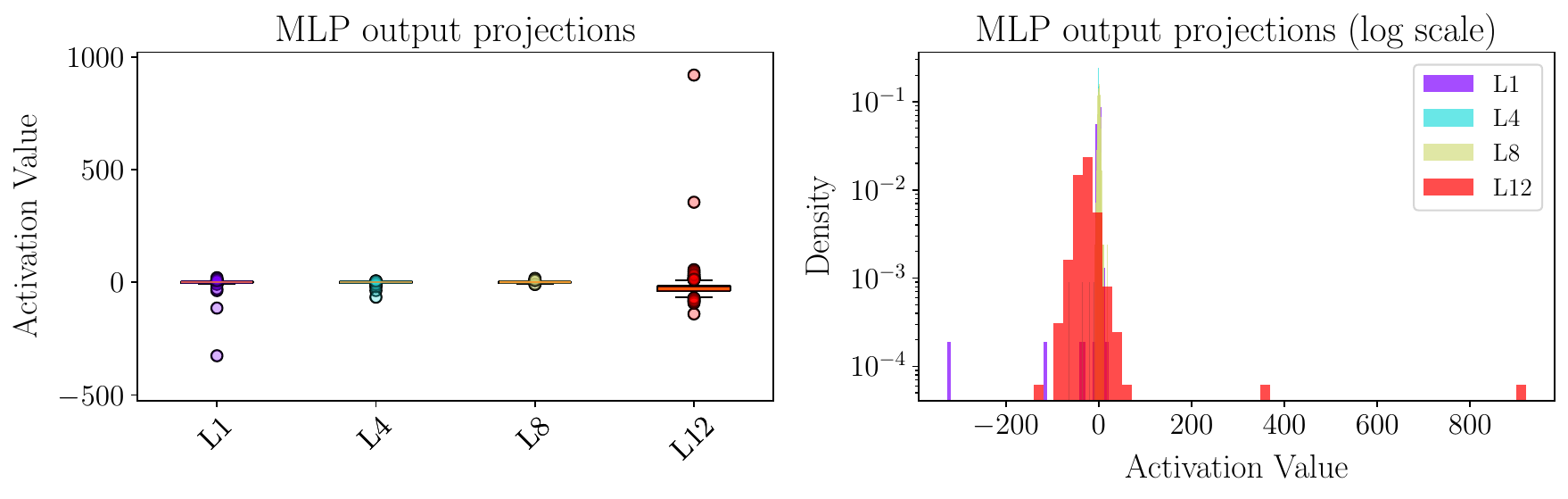}
    \caption{High-magnitude activation outliers and value distribution in the MLP output projection of layers 1, 4, 8, and 12 after a forward pass with the sentence ``model quantization is awesome'', using the 125M-parameter model GPT-Neo \citep{gpt-neo}.}
    \label{fig:transformers-outliers}
\end{figure*}

However, although its premise is simple, quantization is non-trivial, especially for transformers. A challenging aspect of quantizing transformers networks is handling outliers \citep{heo2024rethinking}, which are both frequent and \textit{important} for model predictions \citep{bondarenko2021understanding, bondarenko-do-nothing-2023}. Figure \ref{fig:transformers-outliers} illustrates the activation values of the output projection in the multi-layer perceptron of the model, using a GPT-based model, \textit{gpt-neo-125m}\citep{gpt-neo}. The paper on \textit{ZeroQuant} \citep{yao2022zeroquantefficientaffordableposttraining} provides further discussions on the performance loss of heavily quantizing activations.

By zeroing out or clipping these outliers, performance has been observed to degrade \citep{kovaleva2021bertbustersoutlierdimensions}.
Furthermore, these outliers are found after every linear layer in the query, key, and value projections of transformer networks \citep{dettmers-gptint8} and thus have to be handled during quantization.
Managing these outliers has been attempted to be resolved by many \citep{bondarenko2021understanding, fan2021trainingquantizationnoiseextreme, wei-2022-outlier-supp, wei2023outliersuppressionaccuratequantization, dettmers_case_2023}. Some methods require retraining, and some with modifications to the transformer architecture itself \citep{bondarenko-do-nothing-2023}.
\begin{figure*}[!ht]
    \centering
    \includegraphics[width=1\linewidth]{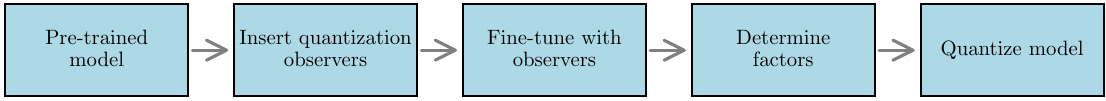}
    \caption{Workflow of quantization-aware training.}
    \label{fig:qat}
\end{figure*}

\begin{figure*}[!ht]
    \centering
    \includegraphics[width=1\linewidth]{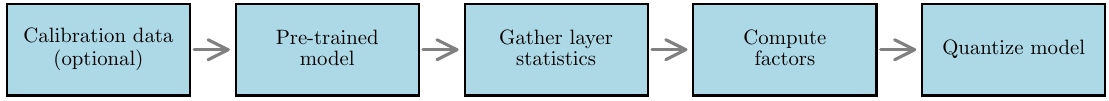}
    \caption{Workflow of post-training quantization.}
    \label{fig:ptq}
\end{figure*}

\paragraph{In-depth Background}
There are two distinct types of quantization: Quantization-Aware Training (QAT) and Post-Training Quantization (PTQ). QAT and PTQ are briefly introduced below, but the following details only consider PTQ, as we are interested in optimizations \textit{after} training a model.
The key steps to both PTQ and QAT are shown in Figures \ref{fig:ptq} and \ref{fig:qat}.

\subsection{QAT}
Quantization-Aware Training (QAT) involves low-precision operations during training by introducing a quantization-dequantization step between the layers, a scheme proposed by \citet{Jacob_2018_CVPR}. During backpropagation, as the quantized function is non-differentiable, the gradients are approximated, e.g., using a \textit{straight-through estimator} \citep{bengio2013estimatingpropagatinggradientsstochastic, yin2019understandingstraightthroughestimatortraining}.
The hypothesis is that the loss function will reduce the negative impact of quantization on performance by optimizing it specifically for low-precision computations. While it may yield higher accuracy, it also requires expensive pre-training.
As of April 2025, a month after the original release, Google published QAT versions of their Gemma-3 series \citep{team2025gemma} of 1B, 4B, 12B and 27B multimodal and multilingual models\footnote{\url{https://huggingface.co/collections/google/gemma-3-qat-67ee61ccacbf2be4195c265b}}, providing powerful models ready to serve on consumer hardware and mobile devices. One can hope this sparks interest for other institutions to train and publish already quantized models.


\subsection{PTQ} Post-training quantization (PTQ) applies a quantization strategy to a pre-trained model, mapping its weights and activations to lower-bit representations.

This mapping can occur in both a linear and non-linear fashion \citep{gobo2020}. However, linear (or \textit{uniform}) quantization is by far the most common approach. This is likely due to how linear operations integrate efficiently with with hardware-optimized matrix operations, e.g., in CUDA and Triton kernels, without additional computations due to non-linear transformations.
The linearity is a mapping between the range of the original values (e.g., FP32) and the target values (e.g., INT8), which depends on two parameters: $s$ (scaling) and $z$ (zero-point, which can be seen as an \textit{offset}). In terms of code, this typically looks like: \texttt{real = scaling\_factor * (quantized - zero\_point)}.

The selection of scaling- and zero-points, linear range selection, and more will be described in subsequent sections, categorized by \textit{asymmetric} and \textit{symmetric} quantization.
First, however, it is beneficial to be aware that just as important a step in quantization is \textit{dequantization}. Dequantization reconstructs an approximation of the original floating-point values and is necessary because neural network layers expect continuous-valued inputs, depending on their training setup. Integers, for example, cannot be directly applied in the operations of a network, so we need to map them back to the original precision.

\paragraph{No ``One size fits all'' Solution}
Quantization is too broad of a term for the many variants we see today, and it is commong to refer to these variants through their quantization of \textit{weights} and \textit{activations} by, e.g., \textit{W8A16} for 8-bit and 16-bit quantization of the weights and activations, respectively. Some also quantize the Key-Value cache, resulting in notations like \textit{W4A8KV4} \citep{lin2024qservew4a8kv4quantizationcodesign}.
An important detail is that if the activations are kept as floating points, the entire computation will rely on floating point arithmetics. All computations will be \textit{integer-based} if the activations are quantized to INT8 or lower.
Integer-based computations benefit from simpler arithmetic, lower memory bandwidth requirements, and improved hardware utilization -- if supported.\footnote{As an example, INT8 support was first added in the Pascal architecture of Nvidia GPUs \citep{harris2016pascal}}
While FP8 has been proposed as a middle ground \citep{micikevicius2022fp8}, its use in quantization has been debated, with inefficiency compared to INT8 for resource-constrained inference workloads \citep{van2023fp8}. However, in the technical report on DeepSeek-V3 \citep{liu2024deepseek}, following the work by \citet{peng2023fp8}, FP8 showed great results for mixed-precision training.

\subsection{Assymetric Quantization}
Asymmetric quantization maps the input range $X_f = [\beta, \alpha]$ that is adjusted around a skewed distribution, useful for activations in neural networks that have non-zero means or are bounded on one side.
This process starts by computing the scaling factor $s$:
\begin{equation}
s = \frac{\alpha - \beta}{2^n - 1}
\end{equation}
\\
where:
$\alpha = \max(X_f)$,
\\
$\beta = \min(X_f)$,
\\
$n$ is the bit-width of the quantized representation (e.g., $n = 8$ for 8-bit quantization).

To properly shift the floating-point range to fit within the integer representation, the zero-point $z$ is computed as:

\begin{equation}
z = \left\lfloor -1 \cdot \frac{\beta}{s} \right\rfloor
\end{equation}
Such that $\beta$ is mapped to the integer value $0$.

\paragraph{Asym. Quantization Function}
All original values $x_f \in X_f$ are quantized as follows:
\begin{equation}
x_q = \text{clamp} \left( \left\lfloor \frac{x_f}{s} \right\rfloor + z , 0, 2^n - 1 \right)
\end{equation}

where
\begin{equation}
\text{clamp}(x; a, b) = \max(a, \min(x, b))
\end{equation}
Ensuring that $x_q$ remains within the valid range of  $[0, 2^n - 1]$.

Note that clamping introduces non-differentiability, which can lead to optimization issues where gradients must be propagated through the quantization process, i.e., in quantization-aware training. Efforts like \textit{SoftClamp} attempt to work around these issues \citep{li2024continuous}.

\paragraph{Asym. Dequantization Function} 
Given a quantized integer value $x_q$, the reconstructed floating-point value $x_f$ is computed as:

\begin{equation}
x_f = s \cdot (x_q - z)
\end{equation}

This transformation maps the discrete quantized values back into the continuous floating-point domain with less precision.

\subsection{Symmetric Quantization}
Symmetric quantization maps the input range $X_f = [-\alpha, \alpha]$ to a zero-centered distribution in lower precision.
The scale factor $s$ is computed as:  
$$
s = \frac{\alpha}{2^{n-1} - 1}
$$
where $\alpha = \max |X_f|$ is the highest absolute floating point value and $n$ is the bit width of the quantized representation.
The denominator ensures a symmetric range, e.g., $[-127, 127]$ for $n=8$.

\paragraph{Sym. Quantization Function}
Each floating-point value \( x_f \in X_f \) is quantized as:  
\[
x_q = \text{clamp} \left( \left\lfloor \frac{x_f}{s} \right\rfloor, - (2^{n-1} - 1), 2^{n-1} - 1 \right)
\]
\\
Unlike asymmetric quantization, symmetric quantization does not introduce a zero-point shift.

\paragraph{Sym. Dequantization Function}
The quantized value $x_q$ is multiplied by the scaling factor:
$
x_f = s \cdot x_q
$
\\
\\
Symmetric quantization works well for weights because they tend to be centered around zero. This is partly due to how weights are initialized (e.g., Glorot or He initialization; \citealp{pmlr-v9-glorot10a, he2015delvingdeeprectifierssurpassing}).
Asymmetric quantization is preferred for activations, as some activation functions, such as ReLU, produce only positive values, leading to a distribution with a non-zero mean. However, some more recently used activation functions like SwiGLU allow small negative values, making their distributions more balanced than ReLU.

\subsection{Static Quantization}
Static quantization relies on a representative calibration dataset, preferably within the same domain as the use-case for the selected model. By observing the intermittent values in the network in this calibration phase, the activation ranges (max and zero-point) and scaling factors can be adjusted accordingly, and are fixed in future inferencing. As all parameters are static, the model can be stored in a fully quantized setting and can benefit from faster inference. For example, an INT8 quantized model relies only on integer kernels, resulting in faster operations with supported hardware.

\subsection{Dynamic Quantization}
Unlike calibration strategies, dynamic quantization performs online calculation of the necessary parameters.
While the weights and biases of a network are fixed\footnote{Biases only influence the new values through addition, and it is less important to quantize the biases than the weights. The majority of speedup lies in matrix multiplication.}, the input ($X$) naturally changes. To perform a dot product with fully quantized values (hence optimizing inference), we must quantize the input and identify the $\alpha$ and $\beta$ values (in the case of asymmetric) and $\alpha$ for symmetric:
\begin{equation}
    y = XW + B
\end{equation}
Dynamic quantization attempts to solve this by running inference for a few different inputs to observe the typical values (min, max) and get a good approximate value for the scale and zero point.
This requires extra overhead, but should perform higher than static as the input range is computed for each $X$.

\subsection{Parameter Selection Strategies}
\label{sec:quant:param}
To get a representative quantized range, we must select parameters ($\alpha$ and $\beta$) that best allow us to quantize- and dequantize back to the original values with minimal errors before finding the scaling- and zero-point parameters.

\subsubsection{Min-max}
A naïve approach is to select the parameters as shown above in describing quantization methods. However, merely selecting min and max will cause large errors in the dequantization step when outliers are present, affecting all values due to our limited number range.

\subsubsection{Percentile}
By simply narrowing the distribution within a percentile, e.g., 99.99\%, before applying min/max, only the outliers will suffer from deviations in the dequantization step. However, as it has been shown that activation outliers are important to the quality of a transformer network \citep{bondarenko2021understanding, bondarenko-do-nothing-2023}, and discarding these may be undesirable. Some choose to keep the outliers in higher precision, which has shown good results \citep{dettmers-gptint8}.

\subsubsection{Mean Squared Error (MSE)}  
MSE minimizes the difference between the original values and their quantized counterparts. This method requires an optimization step to find the optimal scaling parameters that minimize:  

\begin{equation*}  
    \underset{\alpha, \beta}{\mathrm{argmin}} \sum_i (V_i - \hat{V}_i)^2  
\end{equation*}  

where \( V \) represents the original values and \( \hat{V} \) the quantized values.
While this method ensures minimal \textit{average} distortion, it does not account for variable importance.  

\subsubsection{Cross-Entropy}  
When values do not contribute equally to the final output, e.g., after the softmax layer for token probabilities (which we subsequently use for greedy, beam search or top-P selection), cross-entropy (CE) loss can be used instead, which prioritizes to preserve the relative order rather than minimizing reconstruction error:
\begin{equation*}
    \underset{\alpha, \beta}{\mathrm{argmin}} \ \text{CE}(\text{softmax}(V), \text{softmax}(\hat{V}))
\end{equation*}

\subsection{Quantization Granularity}
\begin{figure}
    \centering
    \includegraphics[width=1.0\linewidth]{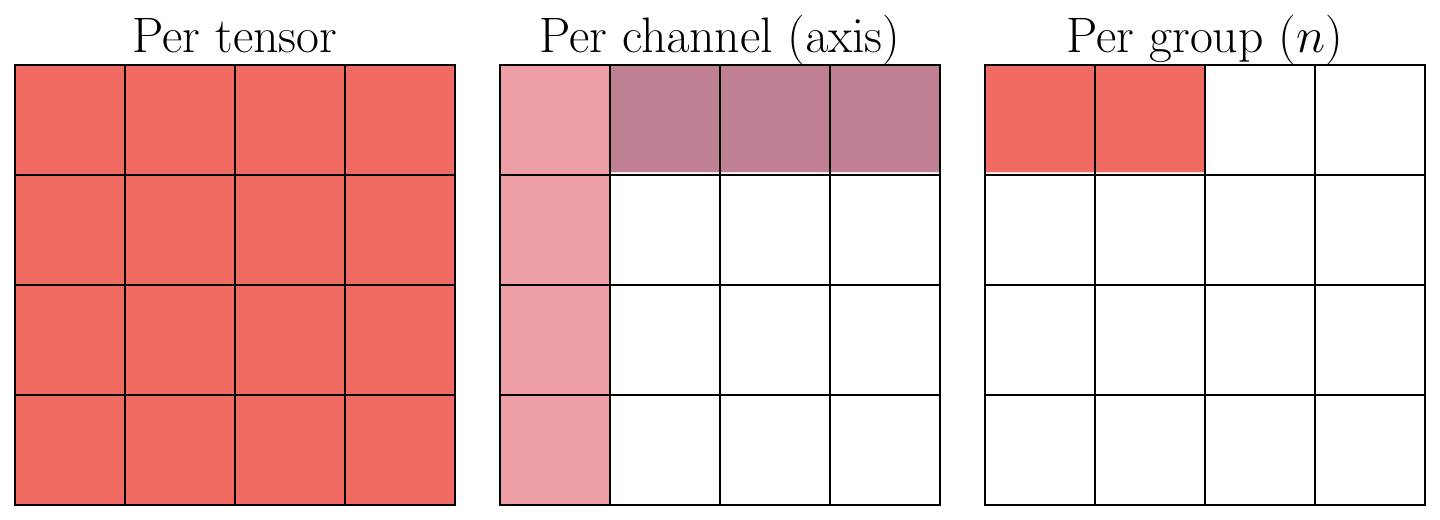}
    \caption{Quantization granularities -- per tensor, per channel (row or column), or by determined group sizes.}
    \label{fig:enter-label}
\end{figure}

The choice of granularity (coarse- to fine-grained) changes the numerical range of original values used in the mapping to quantized values. Constraining the numbers to a certain part of the network, e.g. by changing the \textit{group size}, has shown to increase performance, while slightly increasing overhead (number of effective bits per weight of the final network) and latency \citep{heo2024rethinking}.
From coarse- to fine-grained quantization granularity, the below sections cover quantizing the entire tensor, per channel, and per group.

\subsubsection{Quantizing per Tensor}
Per tensor quantization uses a single scale factor $s$ (and a zero point $z$ if asymmetric) for the entire tensor, following the equations in previous sections.
Although efficient, it may lead to lower accuracy when value distributions vary significantly. 

\begin{figure}
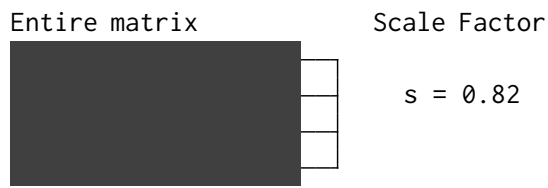

\centering
\begin{verbatim}
Entire matrix              Scale Factor
▓▓▓▓▓▓▓▓──┐
▓▓▓▓▓▓▓▓──┤    s = 0.82
▓▓▓▓▓▓▓▓──┤
▓▓▓▓▓▓▓▓──┘
\end{verbatim}
\caption{Full, single-pass, quantization of a tensor.}
\label{fig:granularity-grouped}
\end{figure}

\subsubsection{Quantizing per Channel}
Per-channel quantization applies separate scaling factors for each output channel in weight tensors, preserving the dynamic range of each channel:

\[
s_c = \frac{\alpha_c}{2^{n-1} - 1}, \quad c = 1,\ldots,C
\]

where $\alpha_c = \max(|X_c|)$ per channel. 
For example, for larger transformer architectures, the value ranges will likely vary between the heads where multi-head attention is used.
Consider
\begin{verbatim}
Head 0 max abs val (alpha): 0.892
Head 1 max abs val (alpha): 0.456
Head 2 max abs val (alpha): 2.341
\end{verbatim}
Considering the entire tensor, the scale parameter would become:
$s_{\text{tensor}} = \frac{2.341}{2^{8-1}-1} \approx 0.0184$
whereas each head would receive higher precision if quantized per channel:
$s_{\text{head}_0} = \frac{0.892}{2^{8-1}-1} \approx 0.0070$.

\begin{figure}
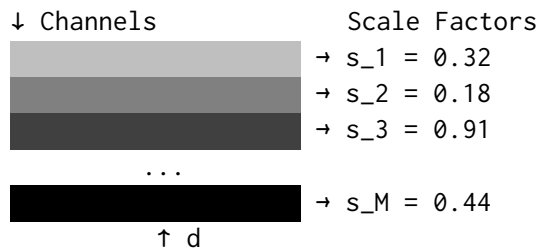

\centering
\begin{verbatim}
↓ Channels             Scale Factors
░░░░░░░░ → s_1 = 0.32
▒▒▒▒▒▒▒▒ → s_2 = 0.18
▓▓▓▓▓▓▓▓ → s_3 = 0.91
         ...
████████ → s_M = 0.44
          ↑ d
\end{verbatim}
\caption{Per-channel, row-wise, quantization, where each line represents a row in a (M,d)-dimensional tensor}
\label{fig:granularity-grouped}
\end{figure}

\subsubsection{Quantizing per Group}
\label{sec:quant:group}
If we go smaller than channel-wise quantization, the $s$ and $z$ parameters comprise significantly more of the total memory required for the final quantized model, although providing better numerical stages. This can be considered both as group-wise and block-wise quantization, depending on the allocation; groups as partitions of rows or columns, or as a \textit{block} or submatrix.
The total \textit{bits per weight} (bpw) comprises the quantized bits + overhead. Consider the following:
\begin{itemize}
    \item Weight dim.: $1024 \cdot 1024 = 1,048,576$
    \item Group size: $g = 128$ weights per group
    \item Total: $\frac{1,048,576}{128} = 8,192$ groups
    \item $\text{bits}_\text{group}$ = $s$ (FP16) + $z$ (INT8, if asym.) $= 24$
\end{itemize}
We can then calculate the overhead per weight, irrespective of the target quantization bit width.
$$
\text{overhead}_{\text{weight}} = \frac{24 \text{ bits}}{128 \text{ weights}}
= 0.1875 \text{ bpw}
$$

\begin{figure}
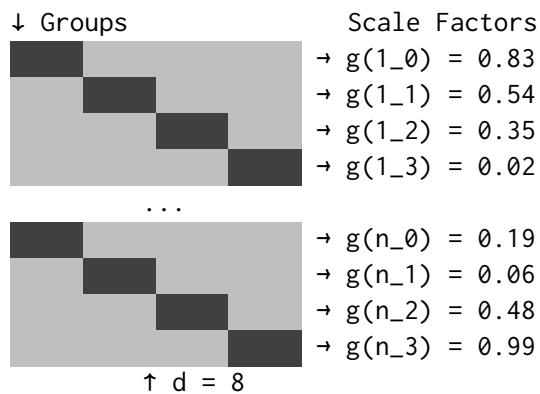

\centering
\begin{verbatim}
↓ Groups               Scale Factors
▓▓░░░░░░ → g(1_0) = 0.83
░░▓▓░░░░ → g(1_1) = 0.54
░░░░▓▓░░ → g(1_2) = 0.35
░░░░░░▓▓ → g(1_3) = 0.02
         ...
▓▓░░░░░░ → g(n_0) = 0.19
░░▓▓░░░░ → g(n_1) = 0.06
░░░░▓▓░░ → g(n_2) = 0.48
░░░░░░▓▓ → g(n_3) = 0.99
         ↑ d = 8
\end{verbatim}
\caption{Example of per group quantization in a sub-row configuration, with group size of 2 in a tensor of dimension $(1,8)$}
\label{fig:granularity-grouped}
\end{figure}


\section{Methods}
\label{sec:app:quantization}
This section presents an overview of post-training quantization methods.
The overview is not exhaustive, but targets well-documented and commonly used methods down to approximately 3 bits per weight. Going beyond the 3-bit threshold has been shown to cause larger deviations from their original weights  \citep{liu2025paretoqscalinglawsextremely}. These observations hold true for both QAT, PTQ, and various fine-tuning approaches.
Details on \textit{extreme} low-bit quantization methods, e.g., down to 1-bit \citep{wang2023bitnet}, can be found in the survey by \citep{gong2024surveylowbitlargelanguage}.
Another important focus is to describe methods supported in common inference libraries, so the end user can implement these for most available open-weight LLMs.
See an overview of methods covered (marked in boldface) in Table \ref{tab:quant-overview} and supported inference libraries.

\begin{table}[ht] 
\centering 
\resizebox{\linewidth}{!}{%
\begin{tabular}{ll}
  \toprule
  \textbf{Library} & \textbf{Quantization} \\ \midrule
  Exllama & \textbf{GPTQ}-based  \\ \midrule
  FastChat & \textbf{AWQ}, \textbf{GPTQ}  \\ \midrule
  gpt-fast & \textbf{GPTQ} (INT4)  \\ \midrule
  llama.cpp & INT2--INT8  \\ \midrule
  LMDeploy & \textbf{AWQ}  \\ \midrule
  OpenVINO & \textbf{AWQ}, \textbf{GPTQ}  \\ \midrule
  ONNX Runtime & \textbf{AWQ}, \textbf{GPTQ}, \textbf{HQQ}  \\ \midrule
  SGLang & \textbf{AWQ}, \textbf{GPTQ}, FP8, INT4  \\ \midrule
  TensorRT-LLM & \textbf{AWQ}, \textbf{SmoothQuant}, FP8, INT8, INT4  \\ \midrule
  Transformers & \textbf{bitsandbytes}, \textbf{GPTQ}, \textbf{HQQ}, +++ \\ \midrule
  vLLM & \textbf{AWQ}, \textbf{GPTQ}, INT8, FP8 \\
  \bottomrule
\end{tabular}
}
\label{tab:quant-overview}
\caption{Overview of popular inferencing libraries and serving engines along with implemented quantization methods. The transformers library is frequently updated with new methods. See the documentation at \url{https://huggingface.co/docs/transformers/en/main_classes/quantization}.
}

\end{table}



\subsection{ZeroQuant}
ZeroQuant \citep{yao_zeroquant-v2_2023} applies quantization to both weights and activations, targeting full model compression across all layers, and was the first method to apply PTQ of multi-billion parameter models.
Supports multiple precision configurations, including mixed-precision weights, but is particularly effective in 8-bit weight and 8-bit activation (W8A8).
They include an optional teacher-student knowledge distillation strategy with reduced memory requirements compared to previous work, improving performance with the mixed-precision schemes set up as W4/8A8 and W4/8A32.
In late 2023, the same research group released an updated version, ZeroQuant-HERO \citep{yao_zeroquant-v2_2023}, building on the original results, focusing specifically on W8A8 quantization.

\paragraph{Availability} The quantization engine is not publicly available.\footnote{\url{https://github.com/deepspeedai/DeepSpeed/issues/2207}}
Parts of the implementations are found in the Deepspeed library on GitHub.\footnote{\url{https://github.com/deepspeedai/DeepSpeed}}


\subsection{LLM.int8() -- bitsandbytes} \citet{dettmers-gptint8} started as an 8-bit PTQ approach, supporting inference on models up to 175B parameters without degradation.
It includes two parts for quantizing:
1) a vector-wise quantization for most weights and activations, and 2) a mixed-precision decomposition, isolating outliers to 16-bit, reducing the effect of high-magnitude outliers on the lower-bit range.
With this scheme, more than 99.9\% of values are quantized in 8 bits, resulting in fast computations for nearly all parts of the network.

\paragraph{Availability} Fully open-sourced on GitHub as the \texttt{bitsandbytes}-library\footnote{\url{https://github.com/bitsandbytes-foundation/bitsandbytes}}, which has been actively developed since, and integrated into the Transformers library \citep{wolf2020huggingfacestransformersstateoftheartnatural} with high-level abstractions for loading models in lower precision. The outlier handling in \texttt{bitsandbytes} has been followed up by GPTQ \citep{frantar2023gptqaccurateposttrainingquantization}, AWQ \citep{lin_awq_2024}, and SmoothQuant \citep{xiao_smoothquant_2024}, among others.

\subsection{GPTQ}
GPTQ \citep{frantar2023gptqaccurateposttrainingquantization} is a one-shot, weight-only PTQ method. It supports 3- and 4-bit quantization, exemplified by running the OPT-175B model \citep{zhang2022optopenpretrainedtransformer} in 3-bit on a single 80GB A100 GPU, and claimed to be the first to perform heavily compressed bit width for these larger models.
GPTQ quantizes weights to minimize the expected increase in loss. Weights associated with high curvature (i.e., large Hessian diagonals) are quantized more conservatively, observing that these weights may reduce model performance.
Authors claim a $3.24\times$ and $4.5\times$ inference speedup over FP16 on an A6000 and A100 GPU, respectively, both with the Ampere architecture.

\paragraph{Availability} Code is available on GitHub. \footnote{\url{https://github.com/IST-DASLab/gptq}}

\subsection{AWQ}
AWQ \citep{lin_awq_2024}, or Activation-aware Weight Quantization, adjusts weight quantization (per-channel) scales based on activation statistics to minimize quantization error.
These statistics are gathered through a calibration dataset, detecting which channels generate outlier features, or \textit{salient weights}.
It supports 4-bit weight quantization and improves accuracy over previous methods on models like LLaMA and OPT.

\paragraph{Availability} The code is available on GitHub.\footnote{\url{https://github.com/mit-han-lab/llm-awq}}
The authors also published a (frequently updated) inference framework named TinyChat.\footnote{\url{https://github.com/mit-han-lab/llm-awq/tree/main/tinychat}}

\begin{figure*}[!ht]
    \centering
    \includegraphics[width=1.0\linewidth]{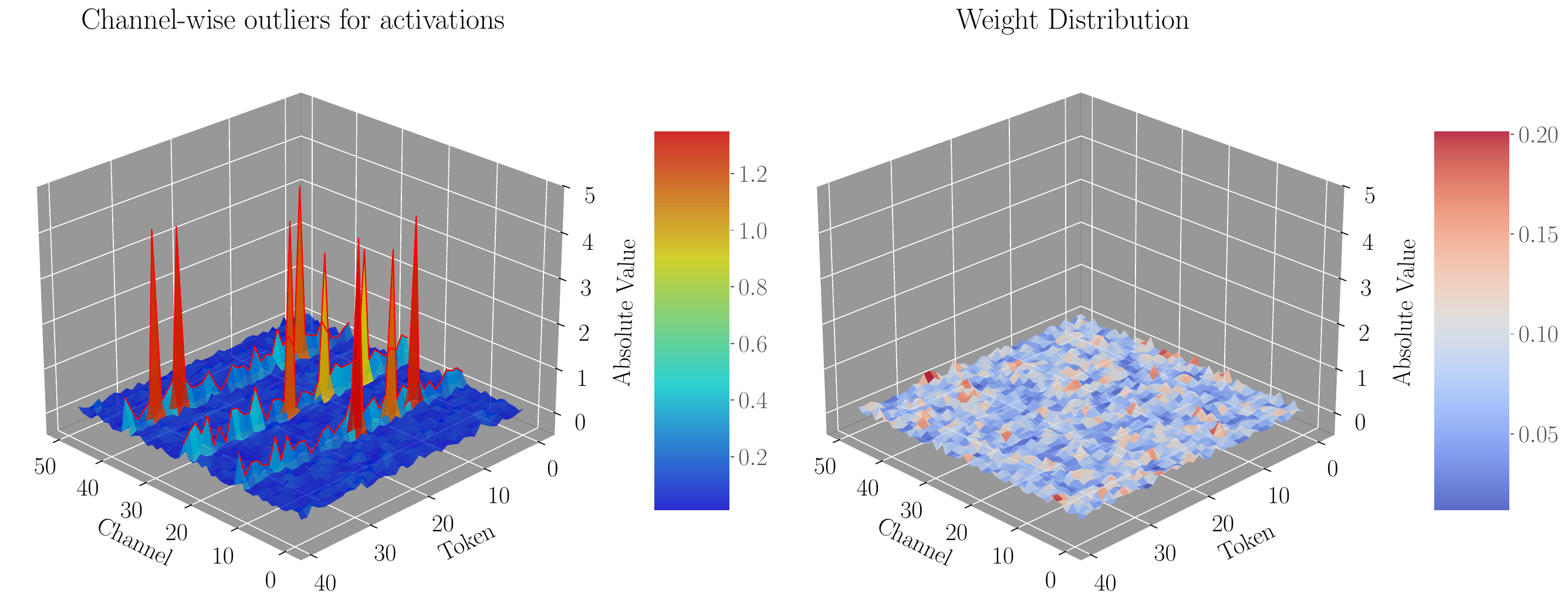}
    \caption{Example showing outliers in certain channels of an activation distribution, compared to a weight distribution. Quantization with spikes (outliers) is a focus area of many PTQ methods.}
    \label{fig:smoothquant}
\end{figure*}

\subsection{SmoothQuant}
SmoothQuant \citep{xiao_smoothquant_2024}, coming from the same lab as AWQ, is designed to improve activation outlier handling, specifically in \textit{W8A8} (8-bit weights and 8-bit activations).
As shown in Figure \ref{fig:smoothquant}, and discussed in earlier sections, activation distributions tend to include sharp spikes in certain channels, especially for models larger than 7B, as noted by \citet{dettmers-gptint8}.
SmoothQuant applies channel-wise scaling to balance the distribution of activations and weights before quantization.
\paragraph{Availability}
Code is open-sourced on GitHub, in great detail.\footnote{\url{https://github.com/mit-han-lab/smoothquant}}







\begin{figure}[!ht]
    \centering
    \includegraphics[width=1.0\linewidth]{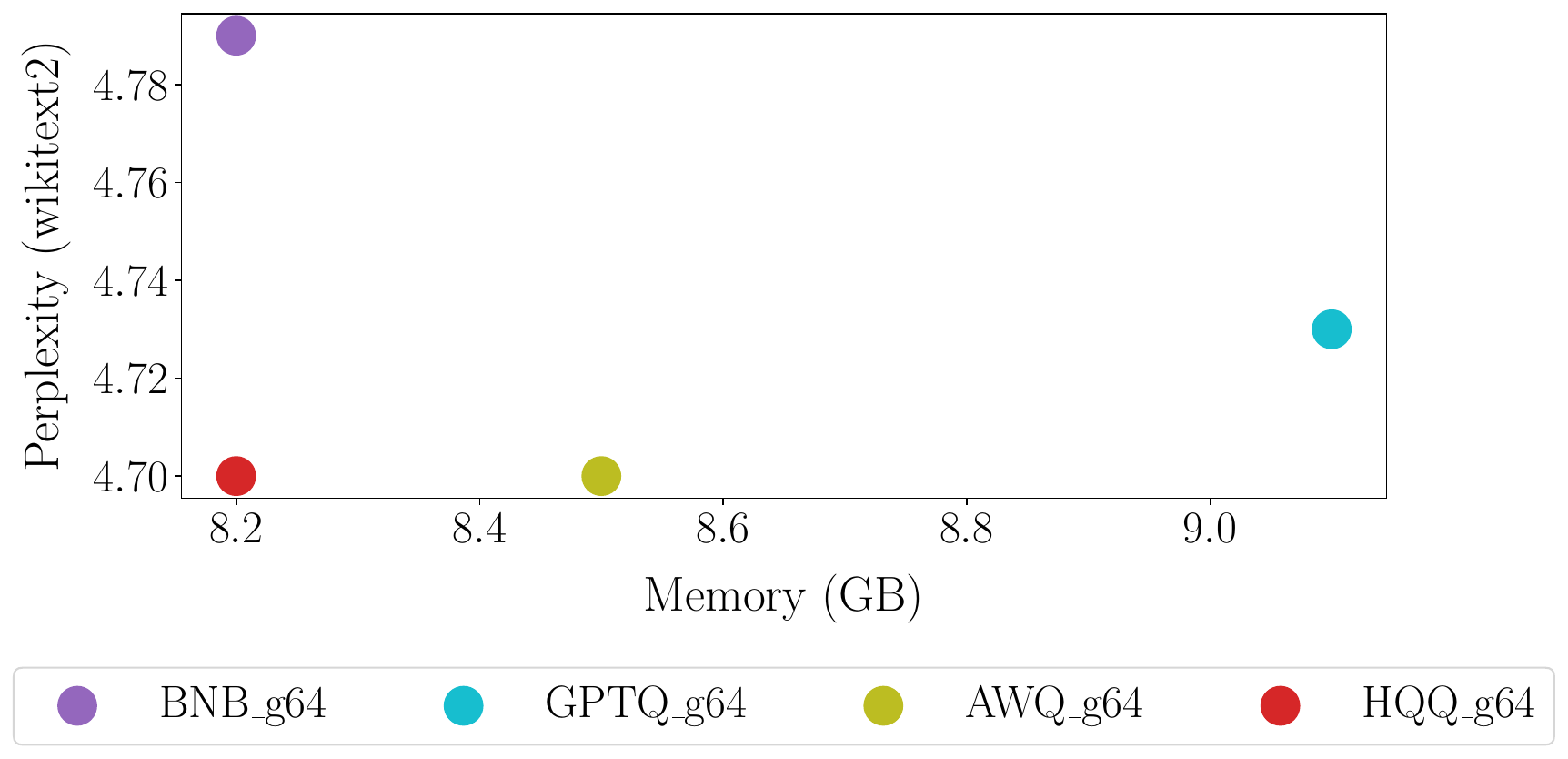}
    \caption{Perplexity on the wikitext2 dataset, as reported by \textit{HQQ}, comparing \texttt{bitsandbytes}, \textit{GPTQ}, and \textit{AWQ} with group size 64 of the \textit{Llama-2 13B} model \citep{touvron2023llama}}
    \label{fig:hqq-quant}
\end{figure}

\subsection{HQQ}
HQQ \citep{badri2023hqq} attempts to solve the problem of relying on calibration data (like GPTQ and AWQ), with a data-free calibration as seen in \texttt{bitsandbytes}\footnote{\url{https://github.com/bitsandbytes-foundation/bitsandbytes}}, avoiding issues like \textit{overfitting} the quantization process towards a smaller subset of calibration data.
HQQ focuses on minimizing quantization error directly in the weight space, handling heavy-tailed weight distributions relying on, e.g., MSE losses as described in Section \ref{sec:quant:param}.
HQQ optimizes, through a Half-Quadratic solver \citep{half-quadratic}, only for the zero-point and keeps the scaling factor static, yielding massive speedups over GPTQ and AWQ with similar memory usage and performance, visualized in Figure \ref{fig:hqq-quant}.

\paragraph{Availability}
Fully open-source and documented, available on GitHub.\footnote{\url{https://github.com/mobiusml/hqq}}

\subsection{Extreme Low-bit Quantization}
Some extreme low-bit quantization methods like DB-LLM \citep{lowbit-chen2024dbllmaccuratedualbinarizationefficient}, PB-LLM \citep{lowbit-shang2023pbllmpartiallybinarizedlarge} and BiLLM \citep{lowbit-huang2024billmpushinglimitposttraining} have shown great results for their size, but remains limited in use for practical model inference \citep{huang2024empirical}.

\subsection{Other Libraries}
\label{sec:app:other}
\paragraph{llama.cpp}
llama.cpp\footnote{\url{https://github.com/ggml-org/llama.cpp}} is a highly popular library, used both standalone and as the backbone in many other \textit{forks} and alternatives to model serving. However, the related \textit{GGUF} format\footnote{\url{https://github.com/ggml-org/ggml/blob/master/docs/gguf.md}}, designed for the tensor library \textit{GGML}\footnote{\url{https://github.com/ggml-org/ggml}} in llama.cpp, lacks official documentation on the implementations of quantization strategies (beyond inspecting specific pull requests). The benefits of llama.cpp is that this format is near architecture-agnostic, running on various GPU- and CPU configurations, and allows offloading specific layers of the model to the CPU/GPU.
The quantization approaches vary between row-wise and block-wise partitioning for scaling- and zero-point factors, along with \textit{super-blocks} composed of several smaller blocks\footnote{\url{https://github.com/ggml-org/llama.cpp/pull/1684}} supporting various configurations of group/block sizes. The naming conventions, commonly Q(bits)\_K\_\{S/M/L\}, refer to the base bitwidth, along with overhead described in Section \ref{sec:quant:group} related to group sizes.

\paragraph{ExLlamaV2}
ExLlamaV2 uses the \textit{EXL2} format for storing the quantized models. Its quantization methods builds on the idea of GPTQ (including calibration data), but where GPTQ typically quantized entire layers or groups in a fixed bitrate (e.g., 4-bit), EXL2 supports mixed-bit quantization and variable group sizes, resulting in the support of any bit rate between 2--8 \textit{bpw}.



\section{Conclusion}
\label{sec:conclusion}
Quantization enables efficient inference of neural networks by reducing the numerical precision of the components of the network. Moreover, it has become the preferred way of reducing size and increasing inference speed of large language models, allowing deployment in resource-constrained settings with reduced hardware- and energy demands
This paper has provided an overview of the foundational principles of quantization, including symmetric/asymmetric approaches, static versus dynamic quantization, parameter selection strategies, and granularity choices.
We have also outlined several PTQ methods and supported inference libraries.

Effective applications of PTQ requires careful considerations, such as dealing with activation outliers without reducing the numerical range for high-quality quantization \citep{kovaleva2021bertbustersoutlierdimensions, wei-2022-outlier-supp, wei2023outliersuppressionaccuratequantization, xiao_smoothquant_2024}.
Moreover, we see a tendency for modern algorithms to support much smaller average \textit{bits per weight} while maintaining surprisingly good performance.

Furthermore, the choice of quantization scheme and granularity, whether per-tensor, per-channel, or per-group/block is intricate, and balances accuracy, computational overhead, and implementation complexity. Certain methods perform the quantization step in minutes \citep{badri2023hqq}, whereas others can require both more resources and time.\citep{egiazarian2024extremecompressionlargelanguage}.
INT4 (or 4-bit) quantization has proven to be a popular choice, and going below 3-bit quantization -- referred to as \textit{extreme low-bit quantization} -- quickly degrades performance \citep{liu2025paretoqscalinglawsextremely}.
There is currently no agreed upon \textit{best practice} for PTQ strategies, and performance across different models (like \textit{Gemma}\citep{gemmateam2024gemma2improvingopen}, Llama \citep{grattafiori2024llama} 

The field of post-training quantization is in active development, and there are several promising areas for future research:

\paragraph{Automated Calibration}
Continue strenghetning calibration methods that automatically determine the optimal quantization configuration (bit-width, granularity, scheme per layer) for a given model and hardware target, along with a specific task.
\paragraph{Data-free Outlier Mitigation}
Improving data-free methods, such as HQQ, for handling outliers for generalizable applications.
\paragraph{Case-specific Evaluation}
We commonly see the evaluations of quantized models across a few select benchmarks, such as perplexity scores. For the end-user, these may not be reflecting the intended purpose. Scores on zero-shot, few-shot, and reasoning tasks may differ greatly, due to their varying importance of instruction-following and more, and there should thus be a focus on fair comparisons in future research.

\bibliographystyle{acl_natbib}
\bibliography{main}

\begin{thebibliography}{75}
\expandafter\ifx\csname natexlab\endcsname\relax\def\natexlab#1{#1}\fi

\bibitem[{Abdin et~al.(2024)Abdin, Aneja, Awadalla, Awadallah, Awan, Bach,
  Bahree, Bakhtiari, Bao, Behl et~al.}]{abdin2024phi}
Marah Abdin, Jyoti Aneja, Hany Awadalla, Ahmed Awadallah, Ammar~Ahmad Awan,
  Nguyen Bach, Amit Bahree, Arash Bakhtiari, Jianmin Bao, Harkirat Behl, et~al.
  2024.
\newblock Phi-3 technical report: A highly capable language model locally on
  your phone.
\newblock \emph{arXiv preprint arXiv:2404.14219}.

\bibitem[{Badri and Shaji(2023)}]{badri2023hqq}
Hicham Badri and Appu Shaji. 2023.
\newblock \href {https://mobiusml.github.io/hqq_blog/} {Half-quadratic
  quantization of large machine learning models}.

\bibitem[{Bai et~al.(2024)Bai, Chai, Ling, Wang, Lu, Zhang, Shi, Yu, Zhu,
  Zhang, Song, Yang, Cheng, and Zhao}]{bai_beyond_2024}
Guangji Bai, Zheng Chai, Chen Ling, Shiyu Wang, Jiaying Lu, Nan Zhang, Tingwei
  Shi, Ziyang Yu, Mengdan Zhu, Yifei Zhang, Xinyuan Song, Carl Yang, Yue Cheng,
  and Liang Zhao. 2024.
\newblock \href {https://doi.org/10.48550/arXiv.2401.00625} {Beyond
  {Efficiency}: {A} {Systematic} {Survey} of {Resource}-{Efficient} {Large}
  {Language} {Models}}.
\newblock ArXiv:2401.00625 [cs].

\bibitem[{Bengio et~al.(2013)Bengio, Léonard, and
  Courville}]{bengio2013estimatingpropagatinggradientsstochastic}
Yoshua Bengio, Nicholas Léonard, and Aaron Courville. 2013.
\newblock \href {http://arxiv.org/abs/1308.3432} {Estimating or propagating
  gradients through stochastic neurons for conditional computation}.

\bibitem[{Black et~al.(2021)Black, Gao, Wang, Leahy, and Biderman}]{gpt-neo}
Sid Black, Leo Gao, Phil Wang, Connor Leahy, and Stella Biderman. 2021.
\newblock \href {https://doi.org/10.5281/zenodo.5297715} {{GPT-Neo: Large Scale
  Autoregressive Language Modeling with Mesh-Tensorflow}}.
\newblock {If you use this software, please cite it using these metadata.}

\bibitem[{Bondarenko et~al.(2021)Bondarenko, Nagel, and
  Blankevoort}]{bondarenko2021understanding}
Yelysei Bondarenko, Markus Nagel, and Tijmen Blankevoort. 2021.
\newblock Understanding and overcoming the challenges of efficient transformer
  quantization.
\newblock \emph{arXiv preprint arXiv:2109.12948}.

\bibitem[{Bondarenko et~al.(2023)Bondarenko, Nagel, and
  Blankevoort}]{bondarenko-do-nothing-2023}
Yelysei Bondarenko, Markus Nagel, and Tijmen Blankevoort. 2023.
\newblock \href
  {https://proceedings.neurips.cc/paper_files/paper/2023/file/edbcb7583fd8921dad78adecfe06a99b-Paper-Conference.pdf}
  {Quantizable transformers: Removing outliers by helping attention heads do
  nothing}.
\newblock In \emph{Advances in Neural Information Processing Systems},
  volume~36, pages 75067--75096. Curran Associates, Inc.

\bibitem[{Chen et~al.(2024)Chen, Lv, Ding, Qin, Zhou, Ding, Liu, Zhang, Guo,
  Liu, and Tao}]{lowbit-chen2024dbllmaccuratedualbinarizationefficient}
Hong Chen, Chengtao Lv, Liang Ding, Haotong Qin, Xiabin Zhou, Yifu Ding, Xuebo
  Liu, Min Zhang, Jinyang Guo, Xianglong Liu, and Dacheng Tao. 2024.
\newblock \href {http://arxiv.org/abs/2402.11960} {Db-llm: Accurate
  dual-binarization for efficient llms}.

\bibitem[{Chitsaz et~al.(2024)Chitsaz, Fournier, Mordido, and
  Chandar}]{chitsaz2024exploringquantizationefficientpretraining}
Kamran Chitsaz, Quentin Fournier, Gonçalo Mordido, and Sarath Chandar. 2024.
\newblock \href {http://arxiv.org/abs/2407.11722} {Exploring quantization for
  efficient pre-training of transformer language models}.

\bibitem[{Courbariaux et~al.(2015)Courbariaux, Bengio, and
  David}]{courbariaux2015trainingdeepneuralnetworks}
Matthieu Courbariaux, Yoshua Bengio, and Jean-Pierre David. 2015.
\newblock \href {http://arxiv.org/abs/1412.7024} {Training deep neural networks
  with low precision multiplications}.

\bibitem[{Dettmers et~al.(2022)Dettmers, Lewis, Belkada, and
  Zettlemoyer}]{dettmers-gptint8}
Tim Dettmers, Mike Lewis, Younes Belkada, and Luke Zettlemoyer. 2022.
\newblock Llm.int8(): 8-bit matrix multiplication for transformers at scale.
\newblock In \emph{Proceedings of the 36th International Conference on Neural
  Information Processing Systems}, NIPS '22, Red Hook, NY, USA. Curran
  Associates Inc.

\bibitem[{Dettmers and Zettlemoyer(2023)}]{dettmers_case_2023}
Tim Dettmers and Luke Zettlemoyer. 2023.
\newblock \href {https://proceedings.mlr.press/v202/dettmers23a.html} {The case
  for 4-bit precision: k-bit {Inference} {Scaling} {Laws}}.
\newblock In \emph{Proceedings of the 40th {International} {Conference} on
  {Machine} {Learning}}, pages 7750--7774. PMLR.
\newblock ISSN: 2640-3498.

\bibitem[{Egiazarian et~al.(2024)Egiazarian, Panferov, Kuznedelev, Frantar,
  Babenko, and Alistarh}]{egiazarian2024extremecompressionlargelanguage}
Vage Egiazarian, Andrei Panferov, Denis Kuznedelev, Elias Frantar, Artem
  Babenko, and Dan Alistarh. 2024.
\newblock \href {http://arxiv.org/abs/2401.06118} {Extreme compression of large
  language models via additive quantization}.

\bibitem[{Fan et~al.(2021)Fan, Stock, Graham, Grave, Gribonval, Jegou, and
  Joulin}]{fan2021trainingquantizationnoiseextreme}
Angela Fan, Pierre Stock, Benjamin Graham, Edouard Grave, Remi Gribonval, Herve
  Jegou, and Armand Joulin. 2021.
\newblock \href {http://arxiv.org/abs/2004.07320} {Training with quantization
  noise for extreme model compression}.

\bibitem[{Farina et~al.(2024)Farina, Ahmad, Taha, Younes, Mesbah, Yu, and
  Pedrycz}]{farina_sparsity_2024}
Mirko Farina, Usman Ahmad, Ahmad Taha, Hussein Younes, Yusuf Mesbah, Xiao Yu,
  and Witold Pedrycz. 2024.
\newblock \href {https://doi.org/10.1016/j.neucom.2024.127468} {Sparsity in
  transformers: {A} systematic literature review}.
\newblock \emph{Neurocomputing}, 582:127468.

\bibitem[{Frantar et~al.(2023)Frantar, Ashkboos, Hoefler, and
  Alistarh}]{frantar2023gptqaccurateposttrainingquantization}
Elias Frantar, Saleh Ashkboos, Torsten Hoefler, and Dan Alistarh. 2023.
\newblock \href {http://arxiv.org/abs/2210.17323} {Gptq: Accurate post-training
  quantization for generative pre-trained transformers}.

\bibitem[{Geman and Reynolds(1992)}]{half-quadratic}
D.~Geman and G.~Reynolds. 1992.
\newblock \href {https://doi.org/10.1109/34.120331} {Constrained restoration
  and the recovery of discontinuities}.
\newblock \emph{IEEE Transactions on Pattern Analysis and Machine
  Intelligence}, 14(3):367--383.

\bibitem[{Gemma~Team(2024)}]{gemmateam2024gemma2improvingopen}
Google~Deepmind Gemma~Team. 2024.
\newblock \href {http://arxiv.org/abs/2408.00118} {Gemma 2: Improving open
  language models at a practical size}.

\bibitem[{Gemma~Team(2025)}]{team2025gemma}
Google~Deepmind Gemma~Team. 2025.
\newblock \href
  {https://storage.googleapis.com/deepmind-media/gemma/Gemma3Report.pdf} {Gemma
  3 technical report}.

\bibitem[{Gholami et~al.(2022)Gholami, Kim, Dong, Yao, Mahoney, and
  Keutzer}]{existing-gholami2022survey}
Amir Gholami, Sehoon Kim, Zhen Dong, Zhewei Yao, Michael~W Mahoney, and Kurt
  Keutzer. 2022.
\newblock A survey of quantization methods for efficient neural network
  inference.
\newblock In \emph{Low-power computer vision}, pages 291--326. Chapman and
  Hall/CRC.

\bibitem[{Glorot and Bengio(2010)}]{pmlr-v9-glorot10a}
Xavier Glorot and Yoshua Bengio. 2010.
\newblock \href {https://proceedings.mlr.press/v9/glorot10a.html}
  {Understanding the difficulty of training deep feedforward neural networks}.
\newblock In \emph{Proceedings of the Thirteenth International Conference on
  Artificial Intelligence and Statistics}, volume~9 of \emph{Proceedings of
  Machine Learning Research}, pages 249--256, Chia Laguna Resort, Sardinia,
  Italy. PMLR.

\bibitem[{Gong et~al.(2024)Gong, Ding, Wang, Lv, Zheng, Du, Qin, Guo, Magno,
  and Liu}]{gong2024surveylowbitlargelanguage}
Ruihao Gong, Yifu Ding, Zining Wang, Chengtao Lv, Xingyu Zheng, Jinyang Du,
  Haotong Qin, Jinyang Guo, Michele Magno, and Xianglong Liu. 2024.
\newblock \href {http://arxiv.org/abs/2409.16694} {A survey of low-bit large
  language models: Basics, systems, and algorithms}.

\bibitem[{Grattafiori et~al.(2024)Grattafiori, Dubey, Jauhri, Pandey, Kadian,
  Al-Dahle, Letman, Mathur, Schelten, Vaughan et~al.}]{grattafiori2024llama}
Aaron Grattafiori, Abhimanyu Dubey, Abhinav Jauhri, Abhinav Pandey, Abhishek
  Kadian, Ahmad Al-Dahle, Aiesha Letman, Akhil Mathur, Alan Schelten, Alex
  Vaughan, et~al. 2024.
\newblock The llama 3 herd of models.
\newblock \emph{arXiv preprint arXiv:2407.21783}.

\bibitem[{Groeneveld et~al.(2024)Groeneveld, Beltagy, Walsh, Bhagia, Kinney,
  Tafjord, Jha, Ivison, Magnusson, Wang, Arora, Atkinson, Authur, Chandu,
  Cohan, Dumas, Elazar, Gu, Hessel, Khot, Merrill, Morrison, Muennighoff, Naik,
  Nam, Peters, Pyatkin, Ravichander, Schwenk, Shah, Smith, Strubell, Subramani,
  Wortsman, Dasigi, Lambert, Richardson, Zettlemoyer, Dodge, Lo, Soldaini,
  Smith, and Hajishirzi}]{groeneveld2024olmoacceleratingsciencelanguage}
Dirk Groeneveld, Iz~Beltagy, Pete Walsh, Akshita Bhagia, Rodney Kinney, Oyvind
  Tafjord, Ananya~Harsh Jha, Hamish Ivison, Ian Magnusson, Yizhong Wang, Shane
  Arora, David Atkinson, Russell Authur, Khyathi~Raghavi Chandu, Arman Cohan,
  Jennifer Dumas, Yanai Elazar, Yuling Gu, Jack Hessel, Tushar Khot, William
  Merrill, Jacob Morrison, Niklas Muennighoff, Aakanksha Naik, Crystal Nam,
  Matthew~E. Peters, Valentina Pyatkin, Abhilasha Ravichander, Dustin Schwenk,
  Saurabh Shah, Will Smith, Emma Strubell, Nishant Subramani, Mitchell
  Wortsman, Pradeep Dasigi, Nathan Lambert, Kyle Richardson, Luke Zettlemoyer,
  Jesse Dodge, Kyle Lo, Luca Soldaini, Noah~A. Smith, and Hannaneh Hajishirzi.
  2024.
\newblock \href {http://arxiv.org/abs/2402.00838} {Olmo: Accelerating the
  science of language models}.

\bibitem[{Gu and Dao(2024)}]{gu2024mambalineartimesequencemodeling}
Albert Gu and Tri Dao. 2024.
\newblock \href {http://arxiv.org/abs/2312.00752} {Mamba: Linear-time sequence
  modeling with selective state spaces}.

\bibitem[{Gu et~al.(2022)Gu, Goel, and
  Ré}]{gu2022efficientlymodelinglongsequences}
Albert Gu, Karan Goel, and Christopher Ré. 2022.
\newblock \href {http://arxiv.org/abs/2111.00396} {Efficiently modeling long
  sequences with structured state spaces}.

\bibitem[{Gunasekar et~al.(2023)Gunasekar, Zhang, Aneja, Mendes, Del~Giorno,
  Gopi, Javaheripi, Kauffmann, de~Rosa, Saarikivi
  et~al.}]{gunasekar2023textbooks}
Suriya Gunasekar, Yi~Zhang, Jyoti Aneja, Caio C{\'e}sar~Teodoro Mendes, Allie
  Del~Giorno, Sivakanth Gopi, Mojan Javaheripi, Piero Kauffmann, Gustavo
  de~Rosa, Olli Saarikivi, et~al. 2023.
\newblock Textbooks are all you need.
\newblock \emph{arXiv preprint arXiv:2306.11644}.

\bibitem[{Gupta et~al.(2015)Gupta, Agrawal, Gopalakrishnan, and
  Narayanan}]{gupta-deep-learning-2015}
Suyog Gupta, Ankur Agrawal, Kailash Gopalakrishnan, and Pritish Narayanan.
  2015.
\newblock \href {https://proceedings.mlr.press/v37/gupta15.html} {Deep learning
  with limited numerical precision}.
\newblock In \emph{Proceedings of the 32nd International Conference on Machine
  Learning}, volume~37 of \emph{Proceedings of Machine Learning Research},
  pages 1737--1746, Lille, France. PMLR.

\bibitem[{Han et~al.(2015)Han, Mao, and Dally}]{han2015deep}
Song Han, Huizi Mao, and William~J Dally. 2015.
\newblock Deep compression: Compressing deep neural networks with pruning,
  trained quantization and huffman coding.
\newblock \emph{arXiv preprint arXiv:1510.00149}.

\bibitem[{Harris(2016)}]{harris2016pascal}
Mark Harris. 2016.
\newblock \href
  {https://developer.nvidia.com/blog/new-pascal-gpus-accelerate-inference-in-the-data-center/}
  {New pascal gpus accelerate inference in the data center}.
\newblock Accessed: 2025-03-19.

\bibitem[{He et~al.(2015)He, Zhang, Ren, and
  Sun}]{he2015delvingdeeprectifierssurpassing}
Kaiming He, Xiangyu Zhang, Shaoqing Ren, and Jian Sun. 2015.
\newblock \href {http://arxiv.org/abs/1502.01852} {Delving deep into
  rectifiers: Surpassing human-level performance on imagenet classification}.

\bibitem[{Heo et~al.(2024)Heo, Kim, Kwon, Kim, Kwon, and
  Lee}]{heo2024rethinking}
Jung~Hwan Heo, Jeonghoon Kim, Beomseok Kwon, Byeongwook Kim, Se~Jung Kwon, and
  Dongsoo Lee. 2024.
\newblock \href {https://openreview.net/forum?id=JzG7kSpjJk} {Rethinking
  channel dimensions to isolate outliers for low-bit weight quantization of
  large language models}.
\newblock In \emph{The Twelfth International Conference on Learning
  Representations}.

\bibitem[{Huang et~al.(2024{\natexlab{a}})Huang, Liu, Qin, Li, Zhang, Liu,
  Magno, and Qi}]{lowbit-huang2024billmpushinglimitposttraining}
Wei Huang, Yangdong Liu, Haotong Qin, Ying Li, Shiming Zhang, Xianglong Liu,
  Michele Magno, and Xiaojuan Qi. 2024{\natexlab{a}}.
\newblock \href {http://arxiv.org/abs/2402.04291} {Billm: Pushing the limit of
  post-training quantization for llms}.

\bibitem[{Huang et~al.(2024{\natexlab{b}})Huang, Zheng, Ma, Qin, Lv, Chen, Luo,
  Qi, Liu, and Magno}]{huang2024empirical}
Wei Huang, Xingyu Zheng, Xudong Ma, Haotong Qin, Chengtao Lv, Hong Chen, Jie
  Luo, Xiaojuan Qi, Xianglong Liu, and Michele Magno. 2024{\natexlab{b}}.
\newblock An empirical study of llama3 quantization: From llms to mllms.
\newblock \emph{Visual Intelligence}, 2(1):36.

\bibitem[{Jacob et~al.(2017)Jacob, Kligys, Chen, Zhu, Tang, Howard, Adam, and
  Kalenichenko}]{jacob2017quantizationtrainingneuralnetworks}
Benoit Jacob, Skirmantas Kligys, Bo~Chen, Menglong Zhu, Matthew Tang, Andrew
  Howard, Hartwig Adam, and Dmitry Kalenichenko. 2017.
\newblock \href {http://arxiv.org/abs/1712.05877} {Quantization and training of
  neural networks for efficient integer-arithmetic-only inference}.

\bibitem[{Jacob et~al.(2018)Jacob, Kligys, Chen, Zhu, Tang, Howard, Adam, and
  Kalenichenko}]{Jacob_2018_CVPR}
Benoit Jacob, Skirmantas Kligys, Bo~Chen, Menglong Zhu, Matthew Tang, Andrew
  Howard, Hartwig Adam, and Dmitry Kalenichenko. 2018.
\newblock Quantization and training of neural networks for efficient
  integer-arithmetic-only inference.
\newblock In \emph{Proceedings of the IEEE Conference on Computer Vision and
  Pattern Recognition (CVPR)}.

\bibitem[{Jiang et~al.(2023)Jiang, Sablayrolles, Mensch, Bamford, Chaplot,
  de~las Casas, Bressand, Lengyel, Lample, Saulnier, Lavaud, Lachaux, Stock,
  Scao, Lavril, Wang, Lacroix, and Sayed}]{jiang2023mistral7b}
Albert~Q. Jiang, Alexandre Sablayrolles, Arthur Mensch, Chris Bamford,
  Devendra~Singh Chaplot, Diego de~las Casas, Florian Bressand, Gianna Lengyel,
  Guillaume Lample, Lucile Saulnier, Lélio~Renard Lavaud, Marie-Anne Lachaux,
  Pierre Stock, Teven~Le Scao, Thibaut Lavril, Thomas Wang, Timothée Lacroix,
  and William~El Sayed. 2023.
\newblock \href {http://arxiv.org/abs/2310.06825} {Mistral 7b}.

\bibitem[{Jin et~al.(2024)Jin, Du, Huang, Liu, Luan, Wang, and
  Xiong}]{jin2024comprehensiveevaluationquantizationstrategies}
Renren Jin, Jiangcun Du, Wuwei Huang, Wei Liu, Jian Luan, Bin Wang, and Deyi
  Xiong. 2024.
\newblock \href {http://arxiv.org/abs/2402.16775} {A comprehensive evaluation
  of quantization strategies for large language models}.

\bibitem[{Kaplan et~al.(2020)Kaplan, McCandlish, Henighan, Brown, Chess, Child,
  Gray, Radford, Wu, and Amodei}]{kaplan2020scalinglawsneurallanguage}
Jared Kaplan, Sam McCandlish, Tom Henighan, Tom~B. Brown, Benjamin Chess, Rewon
  Child, Scott Gray, Alec Radford, Jeffrey Wu, and Dario Amodei. 2020.
\newblock \href {http://arxiv.org/abs/2001.08361} {Scaling laws for neural
  language models}.

\bibitem[{Kovaleva et~al.(2021)Kovaleva, Kulshreshtha, Rogers, and
  Rumshisky}]{kovaleva2021bertbustersoutlierdimensions}
Olga Kovaleva, Saurabh Kulshreshtha, Anna Rogers, and Anna Rumshisky. 2021.
\newblock \href {http://arxiv.org/abs/2105.06990} {Bert busters: Outlier
  dimensions that disrupt transformers}.

\bibitem[{Lang et~al.(2024)Lang, Guo, and
  Huang}]{existing-lang2024comprehensive}
Jiedong Lang, Zhehao Guo, and Shuyu Huang. 2024.
\newblock A comprehensive study on quantization techniques for large language
  models.
\newblock In \emph{2024 4th International Conference on Artificial
  Intelligence, Robotics, and Communication (ICAIRC)}, pages 224--231. IEEE.

\bibitem[{Li et~al.(2024{\natexlab{a}})Li, Hong, Wu, Adbol, and
  Li}]{li2024continuous}
He~Li, Jianhang Hong, Yuanzhuo Wu, Snehal Adbol, and Zonglin Li.
  2024{\natexlab{a}}.
\newblock Continuous approximations for improving quantization aware training
  of llms.
\newblock \emph{arXiv preprint arXiv:2410.10849}.

\bibitem[{Li et~al.(2024{\natexlab{b}})Li, Huang, Chen, Ren, Jiang, Li, Fu, and
  Gao}]{existing-li2024contemporary}
Min Li, Zihao Huang, Lin Chen, Junxing Ren, Miao Jiang, Fengfa Li, Jitao Fu,
  and Chenghua Gao. 2024{\natexlab{b}}.
\newblock Contemporary advances in neural network quantization: A survey.
\newblock In \emph{2024 International Joint Conference on Neural Networks
  (IJCNN)}, pages 1--10. IEEE.

\bibitem[{Lin et~al.(2024{\natexlab{a}})Lin, Tang, Tang, Yang, Chen, Wang,
  Xiao, Dang, Gan, and Han}]{lin_awq_2024}
Ji~Lin, Jiaming Tang, Haotian Tang, Shang Yang, Wei-Ming Chen, Wei-Chen Wang,
  Guangxuan Xiao, Xingyu Dang, Chuang Gan, and Song Han. 2024{\natexlab{a}}.
\newblock \href {https://doi.org/10.48550/arXiv.2306.00978} {{AWQ}:
  {Activation}-aware {Weight} {Quantization} for {LLM} {Compression} and
  {Acceleration}}.
\newblock ArXiv:2306.00978 [cs].

\bibitem[{Lin et~al.(2024{\natexlab{b}})Lin, Tang, Yang, Zhang, Xiao, Gan, and
  Han}]{lin2024qservew4a8kv4quantizationcodesign}
Yujun Lin, Haotian Tang, Shang Yang, Zhekai Zhang, Guangxuan Xiao, Chuang Gan,
  and Song Han. 2024{\natexlab{b}}.
\newblock \href {http://arxiv.org/abs/2405.04532} {Qserve: W4a8kv4 quantization
  and system co-design for efficient llm serving}.

\bibitem[{Liu et~al.(2024)Liu, Feng, Xue, Wang, Wu, Lu, Zhao, Deng, Zhang, Ruan
  et~al.}]{liu2024deepseek}
Aixin Liu, Bei Feng, Bing Xue, Bingxuan Wang, Bochao Wu, Chengda Lu, Chenggang
  Zhao, Chengqi Deng, Chenyu Zhang, Chong Ruan, et~al. 2024.
\newblock Deepseek-v3 technical report.
\newblock \emph{arXiv preprint arXiv:2412.19437}.

\bibitem[{Liu et~al.(2025)Liu, Zhao, Huang, Chen, Zhang, Zhao, Roy, Jin, Xiong,
  Shi, Xiao, Tian, Soran, Krishnamoorthi, Blankevoort, and
  Chandra}]{liu2025paretoqscalinglawsextremely}
Zechun Liu, Changsheng Zhao, Hanxian Huang, Sijia Chen, Jing Zhang, Jiawei
  Zhao, Scott Roy, Lisa Jin, Yunyang Xiong, Yangyang Shi, Lin Xiao, Yuandong
  Tian, Bilge Soran, Raghuraman Krishnamoorthi, Tijmen Blankevoort, and Vikas
  Chandra. 2025.
\newblock \href {http://arxiv.org/abs/2502.02631} {Paretoq: Scaling laws in
  extremely low-bit llm quantization}.

\bibitem[{Ma et~al.(2024)Ma, Wang, Ma, Wang, Wang, Huang, Dong, Wang, Xue, and
  Wei}]{ma2024era}
Shuming Ma, Hongyu Wang, Lingxiao Ma, Lei Wang, Wenhui Wang, Shaohan Huang,
  Lifeng Dong, Ruiping Wang, Jilong Xue, and Furu Wei. 2024.
\newblock The era of 1-bit llms: All large language models are in 1.58 bits.
\newblock \emph{arXiv preprint arXiv:2402.17764}, 1.

\bibitem[{Micikevicius et~al.(2022)Micikevicius, Stosic, Burgess, Cornea,
  Dubey, Grisenthwaite, Ha, Heinecke, Judd, Kamalu
  et~al.}]{micikevicius2022fp8}
Paulius Micikevicius, Dusan Stosic, Neil Burgess, Marius Cornea, Pradeep Dubey,
  Richard Grisenthwaite, Sangwon Ha, Alexander Heinecke, Patrick Judd, John
  Kamalu, et~al. 2022.
\newblock Fp8 formats for deep learning.
\newblock \emph{arXiv preprint arXiv:2209.05433}.

\bibitem[{Nagel et~al.(2021)Nagel, Fournarakis, Amjad, Bondarenko, van Baalen,
  and Blankevoort}]{existing-nagel2021whitepaperneuralnetwork}
Markus Nagel, Marios Fournarakis, Rana~Ali Amjad, Yelysei Bondarenko, Mart van
  Baalen, and Tijmen Blankevoort. 2021.
\newblock \href {http://arxiv.org/abs/2106.08295} {A white paper on neural
  network quantization}.

\bibitem[{OLMo et~al.(2024)OLMo, Walsh, Soldaini, Groeneveld, Lo, Arora,
  Bhagia, Gu, Huang, Jordan, Lambert, Schwenk, Tafjord, Anderson, Atkinson,
  Brahman, Clark, Dasigi, Dziri, Guerquin, Ivison, Koh, Liu, Malik, Merrill,
  Miranda, Morrison, Murray, Nam, Pyatkin, Rangapur, Schmitz, Skjonsberg,
  Wadden, Wilhelm, Wilson, Zettlemoyer, Farhadi, Smith, and
  Hajishirzi}]{olmo20252olmo2furious}
Team OLMo, Pete Walsh, Luca Soldaini, Dirk Groeneveld, Kyle Lo, Shane Arora,
  Akshita Bhagia, Yuling Gu, Shengyi Huang, Matt Jordan, Nathan Lambert, Dustin
  Schwenk, Oyvind Tafjord, Taira Anderson, David Atkinson, Faeze Brahman,
  Christopher Clark, Pradeep Dasigi, Nouha Dziri, Michal Guerquin, Hamish
  Ivison, Pang~Wei Koh, Jiacheng Liu, Saumya Malik, William Merrill, Lester
  James~V. Miranda, Jacob Morrison, Tyler Murray, Crystal Nam, Valentina
  Pyatkin, Aman Rangapur, Michael Schmitz, Sam Skjonsberg, David Wadden,
  Christopher Wilhelm, Michael Wilson, Luke Zettlemoyer, Ali Farhadi, Noah~A.
  Smith, and Hannaneh Hajishirzi. 2024.
\newblock \href {http://arxiv.org/abs/2501.00656} {2 olmo 2 furious}.

\bibitem[{Peng et~al.(2023{\natexlab{a}})Peng, Alcaide, Anthony, Albalak,
  Arcadinho, Biderman, Cao, Cheng, Chung, Grella, GV, He, Hou, Lin, Kazienko,
  Kocon, Kong, Koptyra, Lau, Mantri, Mom, Saito, Song, Tang, Wang, Wind,
  Wozniak, Zhang, Zhang, Zhao, Zhou, Zhou, Zhu, and
  Zhu}]{peng2023rwkvreinventingrnnstransformer}
Bo~Peng, Eric Alcaide, Quentin Anthony, Alon Albalak, Samuel Arcadinho, Stella
  Biderman, Huanqi Cao, Xin Cheng, Michael Chung, Matteo Grella, Kranthi~Kiran
  GV, Xuzheng He, Haowen Hou, Jiaju Lin, Przemyslaw Kazienko, Jan Kocon,
  Jiaming Kong, Bartlomiej Koptyra, Hayden Lau, Krishna Sri~Ipsit Mantri,
  Ferdinand Mom, Atsushi Saito, Guangyu Song, Xiangru Tang, Bolun Wang,
  Johan~S. Wind, Stanislaw Wozniak, Ruichong Zhang, Zhenyuan Zhang, Qihang
  Zhao, Peng Zhou, Qinghua Zhou, Jian Zhu, and Rui-Jie Zhu. 2023{\natexlab{a}}.
\newblock \href {http://arxiv.org/abs/2305.13048} {Rwkv: Reinventing rnns for
  the transformer era}.

\bibitem[{Peng et~al.(2023{\natexlab{b}})Peng, Wu, Wei, Zhao, Yang, Liu, Xiong,
  Yang, Ni, Hu et~al.}]{peng2023fp8}
Houwen Peng, Kan Wu, Yixuan Wei, Guoshuai Zhao, Yuxiang Yang, Ze~Liu, Yifan
  Xiong, Ziyue Yang, Bolin Ni, Jingcheng Hu, et~al. 2023{\natexlab{b}}.
\newblock Fp8-lm: Training fp8 large language models.
\newblock \emph{arXiv preprint arXiv:2310.18313}.

\bibitem[{Poli et~al.(2023)Poli, Massaroli, Nguyen, Fu, Dao, Baccus, Bengio,
  Ermon, and R{\'e}}]{poli2023hyena}
Michael Poli, Stefano Massaroli, Eric Nguyen, Daniel~Y Fu, Tri Dao, Stephen
  Baccus, Yoshua Bengio, Stefano Ermon, and Christopher R{\'e}. 2023.
\newblock Hyena hierarchy: Towards larger convolutional language models.
\newblock In \emph{International Conference on Machine Learning}, pages
  28043--28078. PMLR.

\bibitem[{Shang et~al.(2023)Shang, Yuan, Wu, and
  Dong}]{lowbit-shang2023pbllmpartiallybinarizedlarge}
Yuzhang Shang, Zhihang Yuan, Qiang Wu, and Zhen Dong. 2023.
\newblock \href {http://arxiv.org/abs/2310.00034} {Pb-llm: Partially binarized
  large language models}.

\bibitem[{Shashidhar et~al.(2023)Shashidhar, Chinta, Sahai, Wang, and
  Ji}]{shashidhar2023democratizing}
Sumuk Shashidhar, Abhinav Chinta, Vaibhav Sahai, Zhenhailong Wang, and Heng Ji.
  2023.
\newblock Democratizing llms: An exploration of cost-performance trade-offs in
  self-refined open-source models.
\newblock \emph{arXiv preprint arXiv:2310.07611}.

\bibitem[{Shen et~al.(2024)Shen, Lai, and Li}]{shen-exploring-llm-quant-2024}
Ao~Shen, Zhiquan Lai, and Dongsheng Li. 2024.
\newblock \href {https://doi.org/10.1145/3689236.3695383} {Exploring
  quantization techniques for large-scale language models: Methods, challenges
  and future directions}.
\newblock In \emph{Proceedings of the 2024 9th International Conference on
  Cyber Security and Information Engineering}, ICCSIE '24, page 783–790, New
  York, NY, USA. Association for Computing Machinery.

\bibitem[{Touvron et~al.(2023{\natexlab{a}})Touvron, Lavril, Izacard, Martinet,
  Lachaux, Lacroix, Rozière, Goyal, Hambro, Azhar, Rodriguez, Joulin, Grave,
  and Lample}]{touvron2023llamaopenefficientfoundation}
Hugo Touvron, Thibaut Lavril, Gautier Izacard, Xavier Martinet, Marie-Anne
  Lachaux, Timothée Lacroix, Baptiste Rozière, Naman Goyal, Eric Hambro,
  Faisal Azhar, Aurelien Rodriguez, Armand Joulin, Edouard Grave, and Guillaume
  Lample. 2023{\natexlab{a}}.
\newblock \href {http://arxiv.org/abs/2302.13971} {Llama: Open and efficient
  foundation language models}.

\bibitem[{Touvron et~al.(2023{\natexlab{b}})Touvron, Martin, Stone, Albert,
  Almahairi, Babaei, Bashlykov, Batra, Bhargava, Bhosale
  et~al.}]{touvron2023llama}
Hugo Touvron, Louis Martin, Kevin Stone, Peter Albert, Amjad Almahairi, Yasmine
  Babaei, Nikolay Bashlykov, Soumya Batra, Prajjwal Bhargava, Shruti Bhosale,
  et~al. 2023{\natexlab{b}}.
\newblock Llama 2: Open foundation and fine-tuned chat models.
\newblock \emph{arXiv preprint arXiv:2307.09288}.

\bibitem[{Van~Baalen et~al.(2023)Van~Baalen, Kuzmin, Nair, Ren, Mahurin, Patel,
  Subramanian, Lee, Nagel, Soriaga et~al.}]{van2023fp8}
Mart Van~Baalen, Andrey Kuzmin, Suparna~S Nair, Yuwei Ren, Eric Mahurin, Chirag
  Patel, Sundar Subramanian, Sanghyuk Lee, Markus Nagel, Joseph Soriaga, et~al.
  2023.
\newblock Fp8 versus int8 for efficient deep learning inference.
\newblock \emph{arXiv preprint arXiv:2303.17951}.

\bibitem[{Vaswani et~al.(2017)Vaswani, Shazeer, Parmar, Uszkoreit, Jones,
  Gomez, Kaiser, and Polosukhin}]{vaswani2017attention}
Ashish Vaswani, Noam Shazeer, Niki Parmar, Jakob Uszkoreit, Llion Jones,
  Aidan~N Gomez, {\L}ukasz Kaiser, and Illia Polosukhin. 2017.
\newblock Attention is all you need.
\newblock \emph{Advances in neural information processing systems}, 30.

\bibitem[{Wang et~al.(2024)Wang, Zhang, Zhang, Wu, Mo, Lu, Wang, Li, Xu, Tang,
  He, Ma, Huang, and Wang}]{wang_comprehensive_2024}
Fali Wang, Zhiwei Zhang, Xianren Zhang, Zongyu Wu, Tzuhao Mo, Qiuhao Lu,
  Wanjing Wang, Rui Li, Junjie Xu, Xianfeng Tang, Qi~He, Yao Ma, Ming Huang,
  and Suhang Wang. 2024.
\newblock \href {https://doi.org/10.48550/arXiv.2411.03350} {A {Comprehensive}
  {Survey} of {Small} {Language} {Models} in the {Era} of {Large} {Language}
  {Models}: {Techniques}, {Enhancements}, {Applications}, {Collaboration} with
  {LLMs}, and {Trustworthiness}}.
\newblock ArXiv:2411.03350 [cs].

\bibitem[{Wang et~al.(2023)Wang, Ma, Dong, Huang, Wang, Ma, Yang, Wang, Wu, and
  Wei}]{wang2023bitnet}
Hongyu Wang, Shuming Ma, Li~Dong, Shaohan Huang, Huaijie Wang, Lingxiao Ma, Fan
  Yang, Ruiping Wang, Yi~Wu, and Furu Wei. 2023.
\newblock Bitnet: Scaling 1-bit transformers for large language models.
\newblock \emph{arXiv preprint arXiv:2310.11453}.

\bibitem[{Wei et~al.(2022{\natexlab{a}})Wei, Tay, Bommasani, Raffel, Zoph,
  Borgeaud, Yogatama, Bosma, Zhou, Metzler, Chi, Hashimoto, Vinyals, Liang,
  Dean, and Fedus}]{wei2022emergentabilitieslargelanguage}
Jason Wei, Yi~Tay, Rishi Bommasani, Colin Raffel, Barret Zoph, Sebastian
  Borgeaud, Dani Yogatama, Maarten Bosma, Denny Zhou, Donald Metzler, Ed~H.
  Chi, Tatsunori Hashimoto, Oriol Vinyals, Percy Liang, Jeff Dean, and William
  Fedus. 2022{\natexlab{a}}.
\newblock \href {http://arxiv.org/abs/2206.07682} {Emergent abilities of large
  language models}.

\bibitem[{Wei et~al.(2023)Wei, Zhang, Li, Zhang, Gong, Guo, and
  Liu}]{wei2023outliersuppressionaccuratequantization}
Xiuying Wei, Yunchen Zhang, Yuhang Li, Xiangguo Zhang, Ruihao Gong, Jinyang
  Guo, and Xianglong Liu. 2023.
\newblock \href {http://arxiv.org/abs/2304.09145} {Outlier suppression+:
  Accurate quantization of large language models by equivalent and optimal
  shifting and scaling}.

\bibitem[{Wei et~al.(2022{\natexlab{b}})Wei, Zhang, Zhang, Gong, Zhang, Zhang,
  Yu, and Liu}]{wei-2022-outlier-supp}
Xiuying Wei, Yunchen Zhang, Xiangguo Zhang, Ruihao Gong, Shanghang Zhang,
  Qi~Zhang, Fengwei Yu, and Xianglong Liu. 2022{\natexlab{b}}.
\newblock \href
  {https://proceedings.neurips.cc/paper_files/paper/2022/file/6f6db140de9c9f111b12ef8a216320a9-Paper-Conference.pdf}
  {Outlier suppression: Pushing the limit of low-bit transformer language
  models}.
\newblock In \emph{Advances in Neural Information Processing Systems},
  volume~35, pages 17402--17414. Curran Associates, Inc.

\bibitem[{Wolf et~al.(2020)Wolf, Debut, Sanh, Chaumond, Delangue, Moi, Cistac,
  Rault, Louf, Funtowicz, Davison, Shleifer, von Platen, Ma, Jernite, Plu, Xu,
  Scao, Gugger, Drame, Lhoest, and
  Rush}]{wolf2020huggingfacestransformersstateoftheartnatural}
Thomas Wolf, Lysandre Debut, Victor Sanh, Julien Chaumond, Clement Delangue,
  Anthony Moi, Pierric Cistac, Tim Rault, Rémi Louf, Morgan Funtowicz, Joe
  Davison, Sam Shleifer, Patrick von Platen, Clara Ma, Yacine Jernite, Julien
  Plu, Canwen Xu, Teven~Le Scao, Sylvain Gugger, Mariama Drame, Quentin Lhoest,
  and Alexander~M. Rush. 2020.
\newblock \href {http://arxiv.org/abs/1910.03771} {Huggingface's transformers:
  State-of-the-art natural language processing}.

\bibitem[{Xiao et~al.(2024)Xiao, Lin, Seznec, Wu, Demouth, and
  Han}]{xiao_smoothquant_2024}
Guangxuan Xiao, Ji~Lin, Mickael Seznec, Hao Wu, Julien Demouth, and Song Han.
  2024.
\newblock \href {https://doi.org/10.48550/arXiv.2211.10438} {{SmoothQuant}:
  {Accurate} and {Efficient} {Post}-{Training} {Quantization} for {Large}
  {Language} {Models}}.
\newblock ArXiv:2211.10438 [cs].

\bibitem[{Yao et~al.(2022)Yao, Aminabadi, Zhang, Wu, Li, and
  He}]{yao2022zeroquantefficientaffordableposttraining}
Zhewei Yao, Reza~Yazdani Aminabadi, Minjia Zhang, Xiaoxia Wu, Conglong Li, and
  Yuxiong He. 2022.
\newblock \href {http://arxiv.org/abs/2206.01861} {Zeroquant: Efficient and
  affordable post-training quantization for large-scale transformers}.

\bibitem[{Yao et~al.(2023)Yao, Wu, Li, Youn, and He}]{yao_zeroquant-v2_2023}
Zhewei Yao, Xiaoxia Wu, Cheng Li, Stephen Youn, and Yuxiong He. 2023.
\newblock \href {https://doi.org/10.48550/arXiv.2303.08302} {{ZeroQuant}-{V2}:
  {Exploring} {Post}-training {Quantization} in {LLMs} from {Comprehensive}
  {Study} to {Low} {Rank} {Compensation}}.
\newblock ArXiv:2303.08302 [cs].

\bibitem[{Yin et~al.(2019)Yin, Lyu, Zhang, Osher, Qi, and
  Xin}]{yin2019understandingstraightthroughestimatortraining}
Penghang Yin, Jiancheng Lyu, Shuai Zhang, Stanley Osher, Yingyong Qi, and Jack
  Xin. 2019.
\newblock \href {http://arxiv.org/abs/1903.05662} {Understanding
  straight-through estimator in training activation quantized neural nets}.

\bibitem[{Zadeh et~al.(2020)Zadeh, Edo, Awad, and Moshovos}]{gobo2020}
Ali~Hadi Zadeh, Isak Edo, Omar~Mohamed Awad, and Andreas Moshovos. 2020.
\newblock \href {https://doi.org/10.1109/MICRO50266.2020.00071} {Gobo:
  Quantizing attention-based nlp models for low latency and energy efficient
  inference}.
\newblock In \emph{2020 53rd Annual IEEE/ACM International Symposium on
  Microarchitecture (MICRO)}, pages 811--824.

\bibitem[{Zhang et~al.(2022)Zhang, Roller, Goyal, Artetxe, Chen, Chen, Dewan,
  Diab, Li, Lin, Mihaylov, Ott, Shleifer, Shuster, Simig, Koura, Sridhar, Wang,
  and Zettlemoyer}]{zhang2022optopenpretrainedtransformer}
Susan Zhang, Stephen Roller, Naman Goyal, Mikel Artetxe, Moya Chen, Shuohui
  Chen, Christopher Dewan, Mona Diab, Xian Li, Xi~Victoria Lin, Todor Mihaylov,
  Myle Ott, Sam Shleifer, Kurt Shuster, Daniel Simig, Punit~Singh Koura, Anjali
  Sridhar, Tianlu Wang, and Luke Zettlemoyer. 2022.
\newblock \href {http://arxiv.org/abs/2205.01068} {Opt: Open pre-trained
  transformer language models}.

\bibitem[{Zhou et~al.(2024)Zhou, Ning, Hong, Fu, Xu, Li, Lou, Wang, Yuan, Li,
  Yan, Dai, Zhang, Dong, and Wang}]{zhou_survey_2024}
Zixuan Zhou, Xuefei Ning, Ke~Hong, Tianyu Fu, Jiaming Xu, Shiyao Li, Yuming
  Lou, Luning Wang, Zhihang Yuan, Xiuhong Li, Shengen Yan, Guohao Dai,
  Xiao-Ping Zhang, Yuhan Dong, and Yu~Wang. 2024.
\newblock \href {https://doi.org/10.48550/arXiv.2404.14294} {A {Survey} on
  {Efficient} {Inference} for {Large} {Language} {Models}}.
\newblock ArXiv:2404.14294 [cs].

\bibitem[{Zhu et~al.(2024)Zhu, Li, Liu, Ma, and Wang}]{zhu_survey_2024}
Xunyu Zhu, Jian Li, Yong Liu, Can Ma, and Weiping Wang. 2024.
\newblock \href {https://doi.org/10.48550/arXiv.2308.07633} {A {Survey} on
  {Model} {Compression} for {Large} {Language} {Models}}.
\newblock ArXiv:2308.07633 [cs].

\end{thebibliography}


\end{document}